\documentclass{article}
\usepackage{PRIMEarxiv}
\usepackage[T1]{fontenc}
\usepackage{graphicx}
\usepackage{hyperref}    

\usepackage[utf8]{inputenc}
\usepackage{amsmath}
\usepackage{bbm}
\usepackage{cleveref}
\usepackage{comment}
\usepackage{url}
\usepackage{mathtools}
\usepackage{fancyhdr}
\usepackage{framed}
\usepackage{amsfonts}
\usepackage{amssymb}
\usepackage{enumerate}
\usepackage{color}
\usepackage{caption}
\usepackage{subcaption}
\usepackage[misc]{ifsym}
\usepackage{enumitem}
\usepackage{booktabs}
\usepackage{multirow}
\usepackage{cite}  

\usepackage{bm}
\usepackage{mathtools}
\usepackage{dsfont}

\usepackage{tikz}
\usepackage{mwe}
\usepackage{varwidth}
\usepackage{graphicx,subcaption,ragged2e}
\usepackage{multirow}

\usepackage{tikz}
\usetikzlibrary{arrows.meta, positioning}

\makeatletter
\newcommand{\pfnsymbol}[1]{%
  \textsuperscript{\@fnsymbol{#1}}%
}
\makeatother

\title{Imputation Uncertainty in Interpretable Machine Learning Methods}

\author{
Pegah Golchian$^{1,2}$, Marvin N. Wright$^{1,2}$
\and
$^1$Leibniz Institute for Prevention Research \& Epidemiology – BIPS, Germany
\and
$^2$Faculty of Mathematics and Computer Science, University of Bremen, Germany  \\
\texttt{wright@leibniz-bips.de} \\
}

\begin{document}

\maketitle

\begin{abstract}
In real data, missing values occur frequently, which affects the interpretation with interpretable machine learning (IML) methods. Recent work considers bias and shows that model explanations may differ between imputation methods, while ignoring additional imputation uncertainty and its influence on variance and confidence intervals. We therefore compare the effects of different imputation methods on the confidence interval coverage probabilities of the IML methods permutation feature importance, partial dependence plots and Shapley values. We show that single imputation leads to underestimation of variance and that, in most cases, only multiple imputation is close to nominal coverage.
\end{abstract}
\textbf{Code --- } {\url{https://github.com/bips-hb/iml_imputation_paper}}

\section{Introduction}
As the predictive power of machine learning (ML) models grows, so does the need to explain and interpret them~\cite{goodman2017european,trustworthyAI2019}. Interpretable machine learning (IML) methods provide local as well as global model-agnostic and model-specific explanations~\cite{molnar2022}. These methods are used, for example, in healthcare applications to make the decision-making process more transparent and explainable~\cite{ahmad2018}. However, when working with real data, we often encounter missing values, which can not only weaken the statistical power and prediction performance of a model when not dealt with properly~\cite{van2018flexible,molenberghs2014handbook}, but can also alter their explanations gained through IML methods~\cite{vo2024explainabilitymachinelearningmodels,erez2024impact}. 

Rubin~\cite{rubin1976inference} defines three different missing data patterns: Missing completely at random (MCAR), missing at random (MAR) and missing not at random (MNAR). While with MCAR the probability of a data point being missing is the same for all, with MAR it depends on observed and with MNAR on unobserved data. There are various ways to deal with missing values. Either an ML method can deal with missing values internally, or they have to be handled beforehand. The easiest way is complete-case-analysis, which omits the incomplete data points, but can lead to biased estimates~\cite{van2018flexible}. Another approach is to fill these missing values with single or multiple imputation, where single imputation creates a single imputed dataset and multiple imputation results in several differently imputed datasets. While more sophisticated single imputation methods, e.g. MissForest~\cite{stekhoven2012missforest}, are convenient for providing a single imputed dataset and offering good prediction performance, they do not account for imputation uncertainty. Multiple imputation accounts for this uncertainty by analyzing each imputed dataset and pooling the results with ``Rubin's rules''~\cite{Rubin1987a,white2011multiple} into a point estimate and a standard error. 

Recent work on IML methods with missing data focuses primarily on bias but does not address the effect of imputation methods on the variance and confidence intervals of IML methods (see Section~\ref{sec: related work}). We suggest that the imputation uncertainty should be considered when applying IML methods. To achieve this, we extend the learner uncertainty introduced in Molnar et al.~\cite{molnar2023relating} for IML methods with imputation uncertainty. Our results show that single imputation underestimates the variance, whereas multiple imputation drastically improves the variance estimation. As a result, confidence interval coverage is closer to the nominal coverage level with multiple imputation.

\subsection{Related Work} \label{sec: related work} 
While there is research about assessing uncertainty of explanations, additional imputation uncertainty caused by missing values is not considered~\cite{molnar2023relating,cafri2022understanding,ishwaran2019standard,williamson2020efficient,slack2021reliable}. Several papers notice altering model explanations depending on the chosen way to deal with missing values~\cite{hapfelmeier2014variable,vo2024explainabilitymachinelearningmodels,erez2024impact,shadbahr2023impact,payrovnaziri2021assessing}. Hapfelmeier and Ulm~\cite{hapfelmeier2014variable} study the interpretation quality of variable importance in a random forest by comparing the bias when using multiple imputation, complete case analysis and a self-contained importance measure and do not recommend using complete case analysis. Vo et al.~\cite{vo2024explainabilitymachinelearningmodels} notice strong differences in Shapley values when using different single imputation methods or the internal handling of missing values in XGBoost on some real datasets. Erez et al.~\cite{erez2024impact} address a similar question and compare the effect of different imputation methods on Shapley values on some real datasets, considering the bias and prediction performance. 
None of the aforementioned publications address imputation uncertainty or variance estimation and we are not aware of a study comparing confidence interval coverage probabilities for IML methods with missing data.

\section{Methods} \label{sec: Methods}
In this paper, we consider the imputation uncertainty of global explanation methods that provide overall insights into the relationship between the features and the model prediction or its performance. \emph{Partial dependence (PD) plots}~\cite{friedman2001greedy} visualize the average effect of a specific feature on a model's prediction over grid points. The \emph{permutation feature importance (PFI)}~\cite{breiman2001random,fisher2019all} measures the contribution of each feature by permuting the values and computing the increase of prediction error after permutation. \emph{Shapley values}~\cite{shapley1953} look at the model predictions from a game-theoretical perspective and evaluate the contribution of each feature to the model prediction. Classically, Shapley values are used for local explanations~\cite{lundberg2017} but can also be extended globally~\cite{lundberg2020local,covert2020understanding}. Here we consider the global SHAP feature importance~\cite{molnar2022}, calculated by the mean absolute Shapley values for each feature. 
 
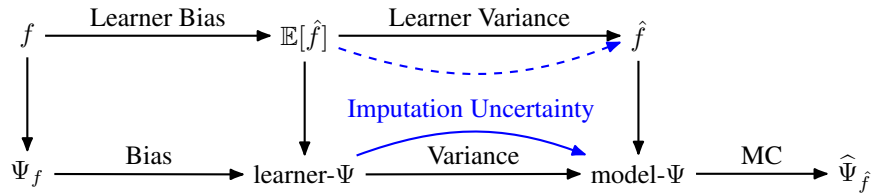
\begin{figure*}[htb]
    \centering
    \begin{tikzpicture}[
        node distance=2.5cm and 3cm, 
        every node/.style={align=center},
        >={Latex[round]},
        every path/.style={draw, thick} 
        ]

        \node (f) {\(f\)};
        \node (Efhat) [right=of f] {\(\mathbb{E}[\hat{f}]\)};
        \node (fhat) [right=of Efhat, xshift=0.75cm] {\(\hat{f}\)};

        \node (IML_f) [below=1.25cm of f] {\(\Psi_{f}\)};
        \node (learner_PFI) [right=of IML_f, below=1.25cm of Efhat] {learner-$\Psi$};
        \node (model_PFI) [right=of learner_PFI, below=1.25cm of fhat] {model-$\Psi$}; 
        \node (PFI_hat) [right=of model_PFI, , xshift=-1.25cm] {\(\widehat{\Psi}_{\hat{f}}\)};
        
        \draw[->] (f) -- node[above] {Learner Bias} (Efhat);
        \draw[->] (Efhat) -- node[above] {Learner Variance} (fhat);

        \draw[->] (IML_f) -- node[above] {Bias} (learner_PFI);
        \draw[->] (learner_PFI) -- node[above] {Variance} (model_PFI);
        \draw[->] (model_PFI) -- node[above] {MC} (PFI_hat); 
        
        \draw[->] (f) -- (IML_f);
        \draw[->] (Efhat) -- (learner_PFI);
        \draw[->] (fhat) -- (model_PFI);

        \draw[->, thick, color=blue] (learner_PFI) to [bend left=20] node[above] {Imputation Uncertainty} (model_PFI);

         \draw[->, dashed, thick, color=blue] (Efhat) to [bend right=20] (fhat);

    \end{tikzpicture}
    \caption{
    The estimates of the model $\hat{f}$ and their explanations with IML methods $\widehat{\Psi}_{\hat{f}}$ (PFI, PD and SHAP) deviate from their ground truth $f$ or $\Psi_{f}$ due to learner bias, variance and Monte Carlo integration (MC). Model-$\Psi$ is the explanation of a fixed model considering Monte Carlo uncertainty, whereas learner-$\Psi$ additionally considers the model uncertainty. Imputation uncertainty as an additional source of error in the model and IML estimates is introduced, extending the illustration from Molnar et al.~\protect\cite{molnar2023relating}.} \label{fig: uncertainty}
\end{figure*}

Molnar et al.~\cite{molnar2023relating} link PD and PFI to the data-generating process (DGP) and consider uncertainty in the interpretation. Related work focuses more on the model uncertainty, i.e., fixing a model and estimating the Monte Carlo uncertainty~\cite{cafri2022understanding,ishwaran2019standard}. For inference about the DGP, this underestimates the variance as shown by Molnar et al.~\cite{molnar2023relating}, who propose the \emph{learner uncertainty} that considers both the Monte Carlo uncertainty and the uncertainty of the model fit, as visualized in Figure~\ref{fig: uncertainty}. 

To create a theoretical foundation that allows statistical comparison, Molnar et al.~\cite{molnar2023relating} define a ground truth version of the PD and PFI that apply the methods to the true function $f$ instead of the trained model $\hat{f}$.
Under the assumption of learner unbiasedness, they show unbiasedness of PD and conditional PFI, and Joseph~\cite{joseph2024interpretabilityinferenceestimationframework} shows asymptotic unbiasedness of Shapley value estimation. 
Here we denote the ground truth of an IML method in general as $\Psi_f: \mathcal{X} \to \mathcal{Z} $ and the explanations on a trained model $\hat{f}$ as $\Psi_{\hat{f}}: \mathcal{X} \to \mathcal{Z}$ that map a set of $p \geq 1$ features $X_S \subset \mathcal{X} \subseteq \mathbb{R}^p$ (e.g. in PFI) or a given feature $x\in X_S$ (e.g. in PD) to their explanations $Z \subset \mathcal{Z}$ for a given model $f$ or $\hat{f}$.
This way, we can describe $\Psi_{\hat{f}}$ as an estimator of the estimand $\Psi_f$.

Let $F$ be the distribution of the models generated by a learning process. We can decompose the error of estimation of the IML methods into bias and learner variance:
\begin{align}
    \mathbb{E}_{F} [ ( \Psi_{\hat{f}} - \Psi_{f} )^2 ] = \operatorname{Bias}^2_F[ \Psi_{\hat{f}}] + \operatorname{Var}_F[ \Psi_{\hat{f}}].
\end{align}

To consider the learner variance, we define the \emph{learner-$\Psi$} as the expected value over $F$, i.e., $\mathbb{E}_{F}[\Psi_{\hat{f}}]$. To estimate $\mathbb{E}_{F}[\Psi_{\hat{f}}]$, we can average the explanations over $k$ different model fits trained on different training data, i.e., by 
\begin{equation}
    \overline{\widehat{\Psi}} = \mathbb{E}_{F}[\Psi_{\hat{f}}] = \frac{1}{k} \sum_{d=1}^k \widehat{\Psi}_{\hat{f}_d}.
\end{equation}

The learner variance can then be estimated as 
\begin{align}
    \widehat{\operatorname{Var}}(\overline{\widehat{\Psi}} ) = \left( \frac{1}{k} +c\right) \cdot \frac{1}{(k-1)} \sum_{d=1}^k (\widehat{\Psi}_{\hat{f}_d} - \overline{\widehat{\Psi}})^2 \label{eq: adj var}
\end{align}
with a correction term $c \in \mathbb{R}$. For an \emph{ideal} setting, where new train/test sets can be drawn from the DGP, we can set $c=0$, which results in the standard variance estimator. With real data, we usually cannot draw repeatedly from the DGP but have to use resampling techniques such as bootstrapping and subsampling instead. These methods provide several training and test datasets, but those are not independent and the standard (naive) variance estimator underestimates the true variance. For this case, Nadeau and Bengio~\cite{nadeau2003inference} recommend setting $c = n_{\text{test}}/n_{\text{train}}$. Even though this correction is only valid under strong assumptions, which are often violated in practice, it improves the variance estimation for the learner uncertainty~\cite{molnar2023relating}. 

Since learner-$\Psi$ are means with estimated variance, the $t$-distribution with $k-1$ degrees of freedom can be used to calculate the confidence bands/intervals
\begin{align}
    \operatorname{CI}_{\overline{\widehat{\Psi}}} = \left[ \overline{\widehat{\Psi}} \pm t_{1-\alpha/2} \sqrt{\widehat{\operatorname{Var}}(\overline{\widehat{\Psi}})}\right]
\end{align}
with the $1-\alpha/2$ quantile $t_{1-\alpha/2}$.
%

%
%
\section{Simulation Studies} \label{sec: Experiments}

\begin{figure*}[!h]
    \centering
    \includegraphics[trim={0.5cm 2.3cm 0.5cm 2.3cm},clip,scale=.45]{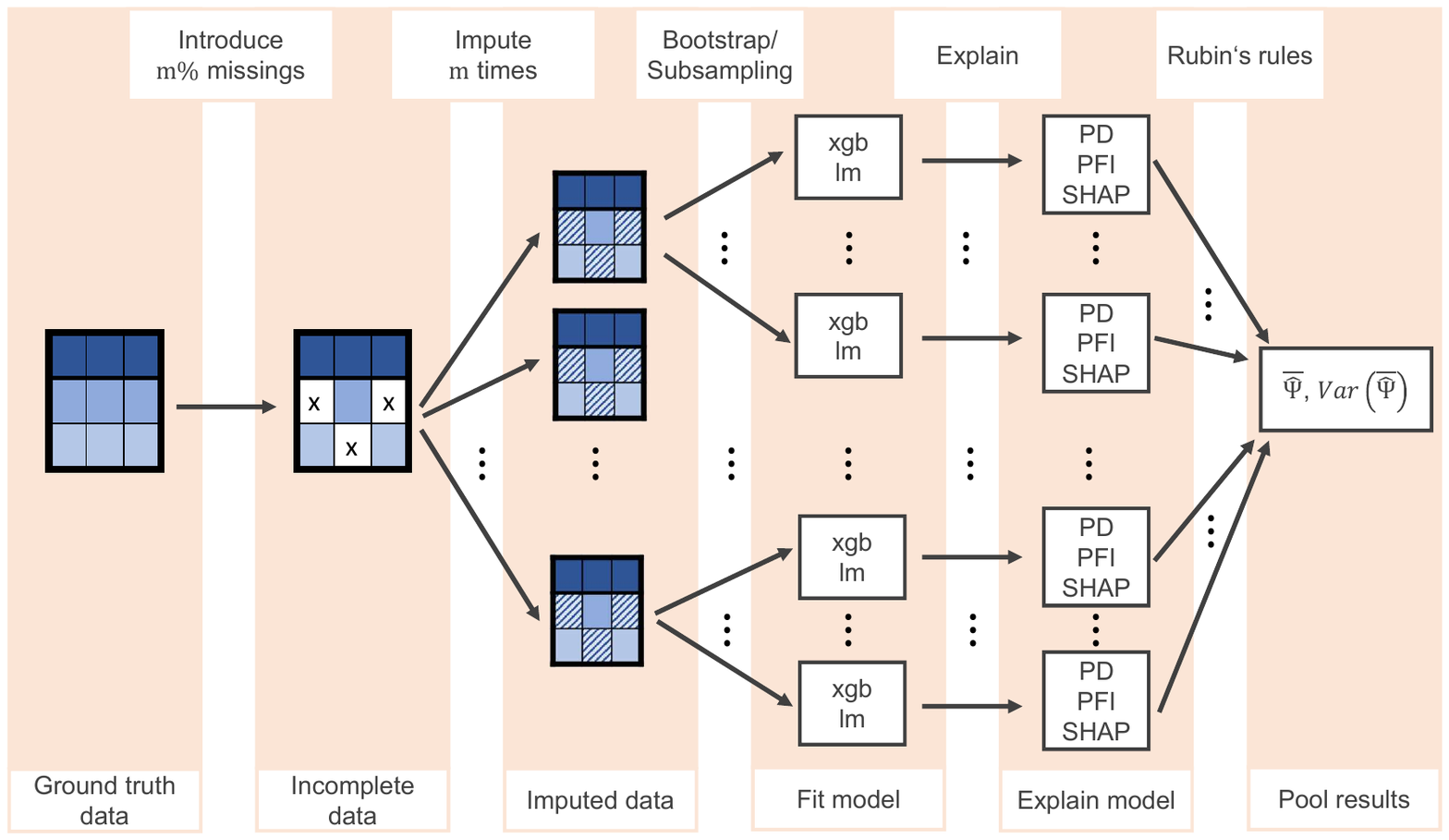}
    \caption{\textbf{Overall procedure of experiment:} 1) Indroduce $m \%$ missing values according to MCAR, MAR and MNAR. 2) Impute $m$ different times. 3) Apply sampling strategy on each imputed dataset, creating $k$ different train and test sets. 4) Fit a model on each of the $m\cdot k$ training datasets. 
    5) Explain each model with an IML method using test data. 6) Pool the results into a point estimate with a standard error using the (un-)adjusted term for the variance (Equation~\ref{eq: adj var}). 7) Repeat the experiment 1000 times. 8) Assess the quality of imputation uncertainty based on coverage, average CI width, and bias.}
    \label{fig: Sketch}
\end{figure*}

\subsection{Setup}
In the following, we compare the confidence interval performance and bias for the learner-PD/PFI/SHAP under the influence of missing values for different settings. While PD and PFI are model-agnostic global methods, due to computational restrictions, we choose global model-specific approaches for SHAP, i.e., TreeSHAP~\cite{lundberg2020local} for XGBoost and exact calculation~\cite{vstrumbelj2014explaining} for the linear regression.

As in Molnar et al.~\cite{molnar2023relating}, we consider a linear DGP, $Y= X_1 - X_2 + \epsilon$ and a non-linear DGP, $Y=X_1-\sqrt{1-X_2} +X_3 X_4 +(X_4/10)^2 +\epsilon$ with $\epsilon \sim \mathcal{N}(0,1)$, where the features are sampled from a multivariate Gaussian distribution $\mathcal{N}(0,\Sigma)$ with Toeplitz covariance matrix $\Sigma_{ij} = 0.5^{|i-j|}$, i.e., adjacent features are mostly correlated. We sample $n=1000$ data points from the chosen DGP and simulate missing values according to MCAR, MAR and MNAR in the resulting dataset with missingness proportions of 0.1, 0.2, and 0.4. To simulate MAR and MNAR, we choose a ranking mechanism~\cite{missMethods_R}, where for MAR it depends on randomly chosen observed features and for MNAR on the features themselves\footnote{For MAR, we have missings in half of the features to comply with the MAR definition. For MCAR and MNAR we have missings in all features.}. These missing values are then imputed with single imputation and multiple imputation. For single imputation, we use mean imputation and MissForest~\cite{missranger}, for multiple imputation \emph{multivariate imputation by chained equations} (MICE)~\cite{mice}, once with predictive mean matching (MICE PMM) and with random forests (MICE RF). 

On each imputed dataset, we apply bootstrap and subsampling (both with 20 resampling iterations), fit a linear model (lm) and an XGBoost (xgb)~\cite{xgboost_chen2016} model\footnote{With maximum 20 boosting iterations and a maximum depth of a tree of 2.} on each resampled dataset and estimate the PD, PFI and SHAP on the respective test data. From these 0-20 model refits, we calculate the learner-PD/PFI/SHAP and their respective confidence intervals. For each setting, we repeat this experiment 1000 times and calculate the confidence interval (CI) coverage probability for a nominal rate of 0.95, i.e., how often the estimated CI covers the expected learner-PD/PFI/SHAP, which is calculated by averaging over 10,000 replications of the experiment without missing values. Furthermore, the average width of the CI and bias are examined.

For both the \emph{bootstrap} and \emph{subsampling} approach, we compare adjusted and unadjusted variance estimation as described above~\cite{nadeau2003inference}\footnote{With missing values, we cannot consider an \emph{ideal} setting as in Molnar et al.~\cite{molnar2023relating} (where we repeatedly sample new data from the DGP for variance estimation) because this would require a separate imputation for each sample from the DGP.}. For multiple imputation, we increase the number of multiple imputations with the missingness proportion and impute 10, 20 and 40 times for missingness proportions of 0.1, 0.2, and 0.4, respectively, as recommended by White et al.~\cite{white2011multiple}. We follow the \emph{MI Boot} approach as recommended by Schomaker and Heumann~\cite{schomaker2018bootstrap}, i.e., we apply the bootstrap (and subsampling) to each imputed dataset. We then use these to fit models, calculate the explanations with the IML methods, estimate variances and pool them by ``Rubin's rules''~\cite{Rubin1987a,white2011multiple} into point estimates with standard errors and corresponding confidence intervals. The overall procedure is visualized in Figure~\ref{fig: Sketch}.

\subsection{Results}
\begin{figure*}[!h]
    \centering
    \includegraphics[width=0.8\textwidth]{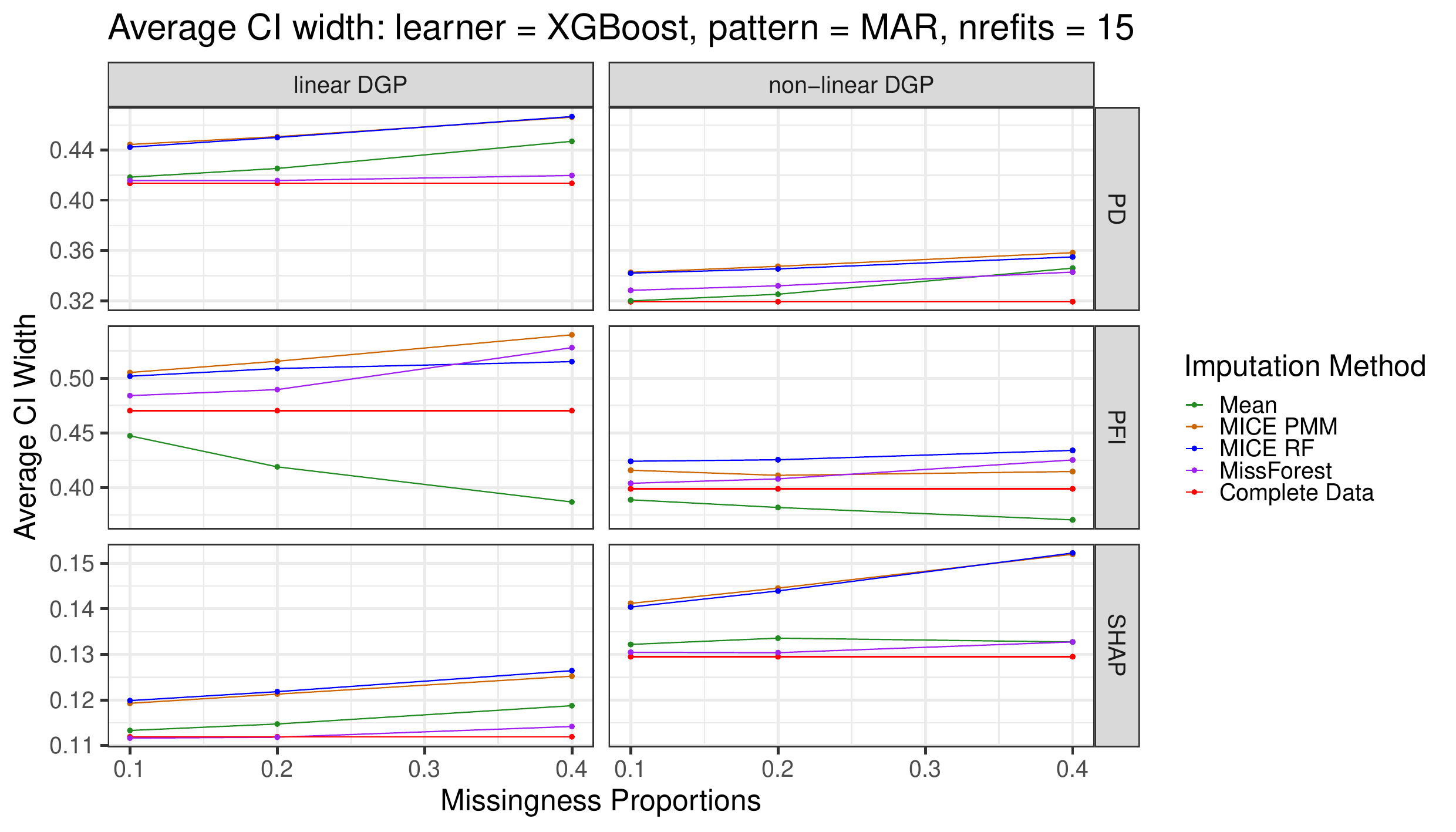}
    \caption{Average CI width across missingness proportions of bootstrapped XGBoost. The models were refitted 15 times. Missingness was introduced under a MAR pattern and imputed using various methods, compared to the ground truth from the complete dataset (red). Results are averaged over 1000 replicates.}
    \label{fig: Sim_example width}
\end{figure*}

In the following, we show simulation results for all the IML methods with the XGBoost learner, a MAR pattern and the bootstrap approach. Figure~\ref{fig: Sim_example cov} displays coverage probabilities with a missingness rate of 0.4, whereas Figure~\ref{fig: Sim_example width} shows the average width of the confidence intervals and Figure~\ref{fig: Sim_example bias} the bias over all missingness proportions and 15 refits. Additional results for bootstrap and subsampling can be found in supplementary figures in the corresponding GitHub repository \url{https://github.com/bips-hb/iml_imputation_paper}.
%

\subsubsection{Coverage}
\begin{figure*}[t]
    \centering
    \includegraphics[width=0.8\textwidth]{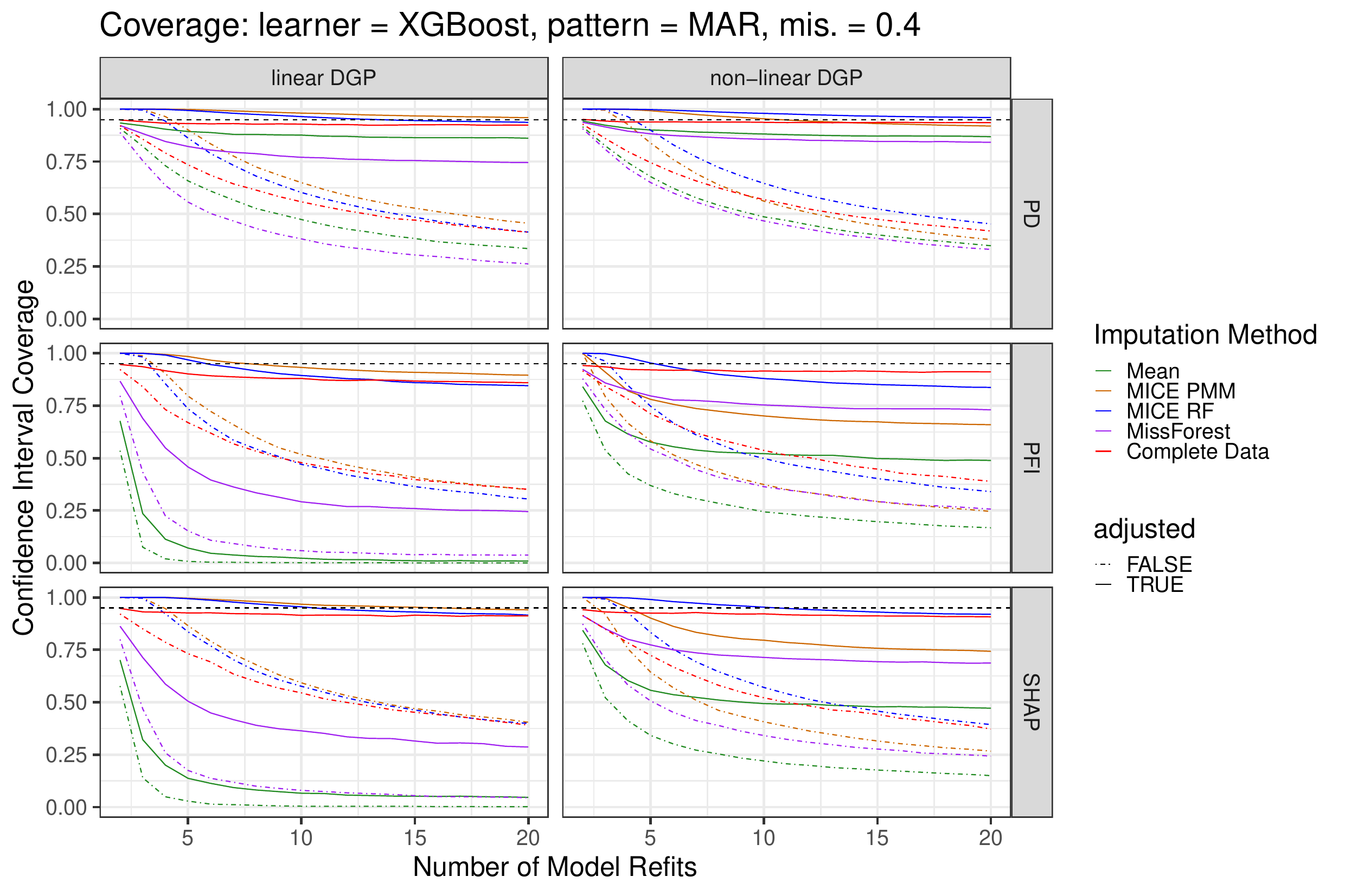}
    \caption{Coverage rates across the number of model refits of bootstrapped XGBoost. 40\% missingness was introduced under a MAR pattern and imputed using various methods, compared to the ground truth from the complete dataset (red). Results are averaged over 1000 replicates. The black dashed line indicates the nominal coverage level of 0.95.}
    \label{fig: Sim_example cov}
\end{figure*}

Figure~\ref{fig: Sim_example cov} shows that variance estimation without adjustment generally leads to poor coverage, as already reported by Molnar et al.~\cite{molnar2023relating} for complete data. Comparing the imputation approaches, MICE tends to have higher coverage for the linear DGP than the full dataset, with MICE PMM often being better than MICE RF. MissForest and mean imputation perform worse for SHAP and PFI, with mean imputation exhibiting particularly poor coverage. They perform better for PD, but still poorly, with mean imputation performing slightly better than MissForest. For the non-linear DGP, MissForest and mean imputation have better coverage compared to the linear one, but remain poor. Except for PFI, MICE RF achieves higher coverage than the complete dataset. MICE PMM also achieves good coverage for PD, but is worse for SHAP and PFI.

When comparing different missingness rates (see Figures \ref{fig: cov pdp}-\ref{fig: cov shap} in the appendix) for XGBoost, the coverage generally decreases along with the missingness. The coverage rate of the MICE methods remains relatively stable, in particular, for MICE PMM in linear DGP and MICE RF for non-linear DGP. While the coverage rate of MissForest is close to the complete data case for a missingness proportion of 0.1, the coverage of mean imputation is considerably lower (except for PD). Over the missing patterns, the plots are very similar, though the coverage rate can get poorer at MNAR.

For the linear model learner (see Figures \ref{fig: cov pdp}-\ref{fig: cov shap} in the appendix), the behavior is quite similar to XGBoost for SHAP and PFI, with the difference that for the linear DGP, the coverage decreases more for MICE RF and mean imputation with the missingness rate. In contrast, for the non-linear DGP the coverage of the methods tends to be slightly closer to the nominal rate than XGBoost. For MCAR and MNAR, MICE PMM has as good a coverage as MICE RF. For PD, MissForest and mean imputation lead to poorer coverage, particularly for the linear DGP. However, MissForest performs as well as or better than mean imputation (except for MCAR and non-linear DGP). For MNAR, the performance of all methods is considerably worse. 

The subsampling approach generally leads to similar behavior as bootstrapping, with in most cases equal or lower coverage.

\subsubsection{Average CI Width}
Provided there is acceptable coverage, we would like the average width of the confidence intervals to be as small as possible. 
Based on the literature, we expect this width to get smaller over the number of refits, which is what our results show\footnote{In GitHub repository: \url{https://github.com/bips-hb/iml_imputation_paper}.}. In the following, we set the number of refits to 15 as recommended in the literature~\cite{molnar2023relating,nadeau2003inference} and describe how the CI width changes across the missingness proportions.

Figure~\ref{fig: Sim_example width} shows that in general the interval width increases with the missingness proportion, except for mean imputation for PFI where it decreases. Both MICE methods generally have wider confidence intervals than the other methods, followed by mean imputation for PD and SHAP and MissForest for PFI. The average width for MissForest is close to the ground truth for a missingness proportion of 0.1 and except for PFI has the smallest increase. The width of the non-linear DGP is smaller for PFI and PD than that of the linear DGP and slightly larger for SHAP. Across the missing data patterns (see Figures \ref{fig: aw pdp}-\ref{fig: aw shap} in the appendix), the trend stays similar.

For the linear model learner (see Figures \ref{fig: aw pdp}-\ref{fig: aw shap} in the appendix), the CI widths are generally slightly larger compared to XGBoost for PFI and SHAP and smaller for PD. Regarding the linear DGP, there is a similar trend as with XGBoost, with the difference that the width of mean imputation decreases with the missingness proportions for SHAP. For PFI, MissForest has the same CI width as MICE at 0.1 and gets larger than MICE for 0.4 missingness. For PD, mean imputation generally has the smallest width in contrast to XGBoost. For the non-linear DGP and SHAP, the width with MissForest increases more than before. In the case of PFI, the width does not decrease as much for the mean imputation and increases for MICE RF, MissForest and MICE PMM, in ascending order, different from before. In contrast to XGBoost, for PD, MissForest has the smallest width and the width is generally slightly larger for the non-linear DGP than the linear DGP. Across the missing data patterns, the trend stays similar except for PD for the linear DGP, where the CI width of mean imputation increases with missingness for MCAR, stays stable for MAR and decreases for MNAR.

Subsampling shows mostly a similar pattern to bootstrap, although with a slightly smaller width.

\subsubsection{Bias} 
As for the average width, we describe the change of the bias over the missingness rate for a fixed number of 15 refits. Since we average over the 1000 repetitions for each refit, the mean of the bias is almost the same over all refits. 

\begin{figure*}
    \centering
    \includegraphics[width=0.8\textwidth]{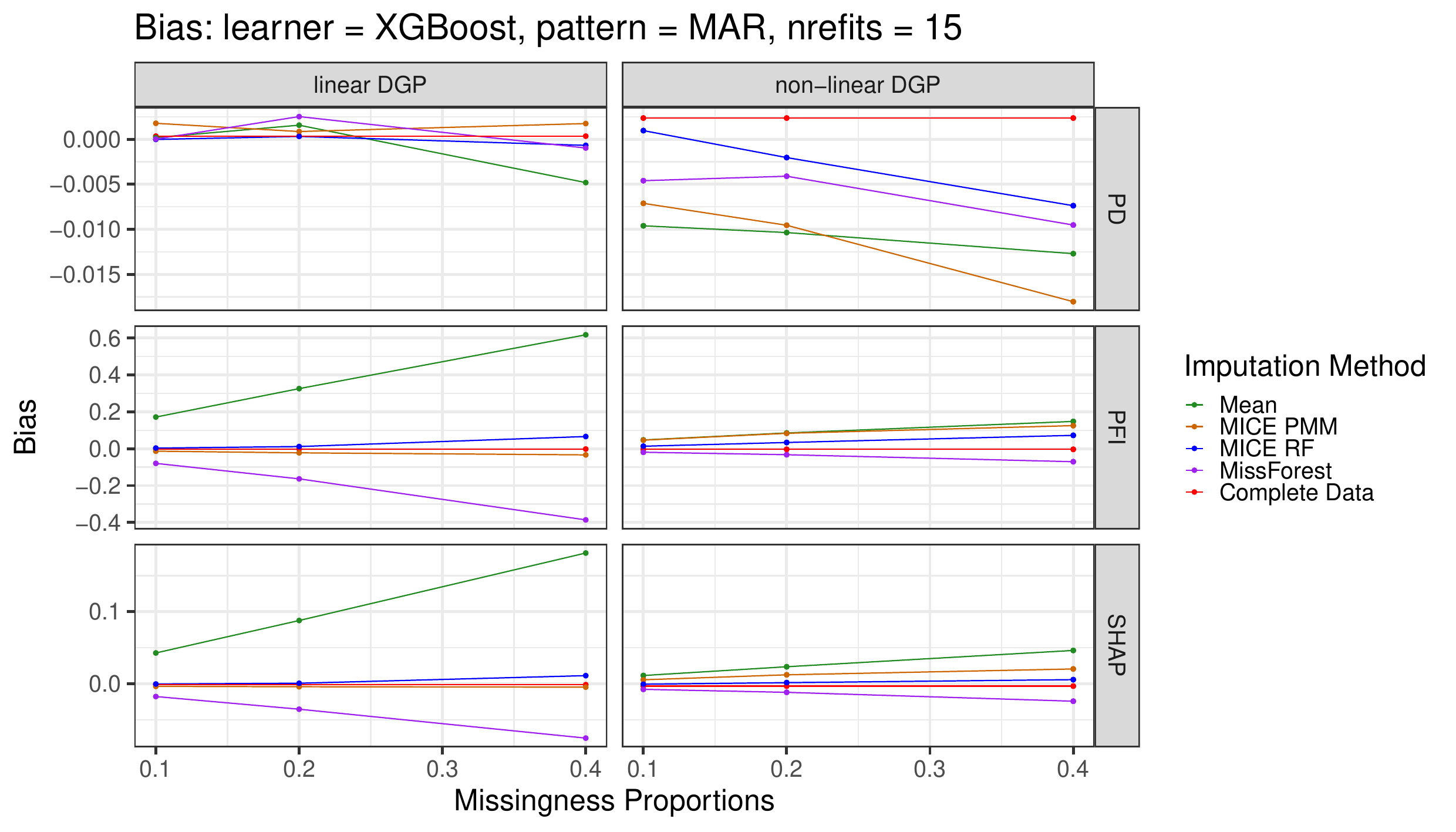}
    \caption{Bias across missingness proportions of bootstrapped XGBoost. The models were refitted 15 times. Missingness was introduced under a MAR pattern and imputed using various methods. Results are averaged over 1000 replicates.}
%
    \label{fig: Sim_example bias}
\end{figure*}

Figure~\ref{fig: Sim_example bias} shows that the bias for PD is very small across all imputation methods, and grows slightly in a negative direction for a non-linear DGP. Here, MICE RF has the smallest bias. For PFI and SHAP for the linear DGP, mean imputation has the largest positive bias, which increases strongly with the missingness rate. This is followed by MissForest, which has an increasing negative bias. The bias for MICE RF grows to a much lesser extent and is very close to zero, whereas the bias for MICE PMM is also very close to zero but with a negative tendency. Considering the different scale for the non-linear DGP, the biases are generally smaller. For PFI, MissForest and MICE RF have the smallest bias. For SHAP, MICE RF has the smallest bias, and MICE PMM and MissForest share the second place. The behavior is similar for the different missing data patterns (see Figures \ref{fig: bias pdp}-\ref{fig: bias shap} in the appendix). 

For the linear model learner, the bias for PD is again very small for both DGP's. Regarding PFI and SHAP for the linear DGP, we observe the same pattern over all settings with a slightly larger bias. MICE PMM is also very close to zero, but with a tendency towards a positive bias. For the non-linear DGP, the mean imputation performs well, is smaller than or equal to MissForest. For SHAP, the bias of MICE RF remains small, while MICE PMM has a negative bias, but is always smaller than MissForest. In contrast to XGBoost, for PFI, MICE PMM has almost no bias and performs equally well or better than MICE RF.

The results for subsampling show a similar pattern to the bootstrap approach.

\section{Real Data Example} \label{sec: example}
The wine dataset from the UCI repository~\cite{wine_quality_186,CorCer09} aims to predict the quality (on a scale from 0 to 10) of red variants of the Portuguese vinho verde based on physicochemical tests. The dataset has 1599 observations of 12 numerical variables with no missing values. We model the regression task with XGBoost with a maximum of 20 boosting iterations and a maximum tree depth of 2 using a bootstrap approach. Now we are interested in interpreting the dataset with a PD plot, PFI and global SHAP feature importance. To what extent would the interpretation change if we had missing values and used imputation methods?

As recommended in Molnar et al.~\cite{molnar2023relating}, we consider the learner uncertainty in the interpretation and calculate confidence intervals over 15 model refits. We simulate 40 \% missing data according to an MCAR pattern~\cite{missMethods_R} and impute with mean and MICE PMM~\cite{mice} with 40 multiple imputations, as recommended by White et al.~\cite{white2011multiple}. We use an adjusted bootstrap~\cite{nadeau2003inference} approach on the imputed dataset~\cite{schomaker2018bootstrap}. The mean squared error averaged over the refits is 0.41 with complete data, 0.48 with mean imputation and 0.42 with MICE. While the mean performance over the refits is not much affected by the imputation, the interpretation is.

\begin{figure*}
    \centering
    \includegraphics[width=0.9\textwidth]{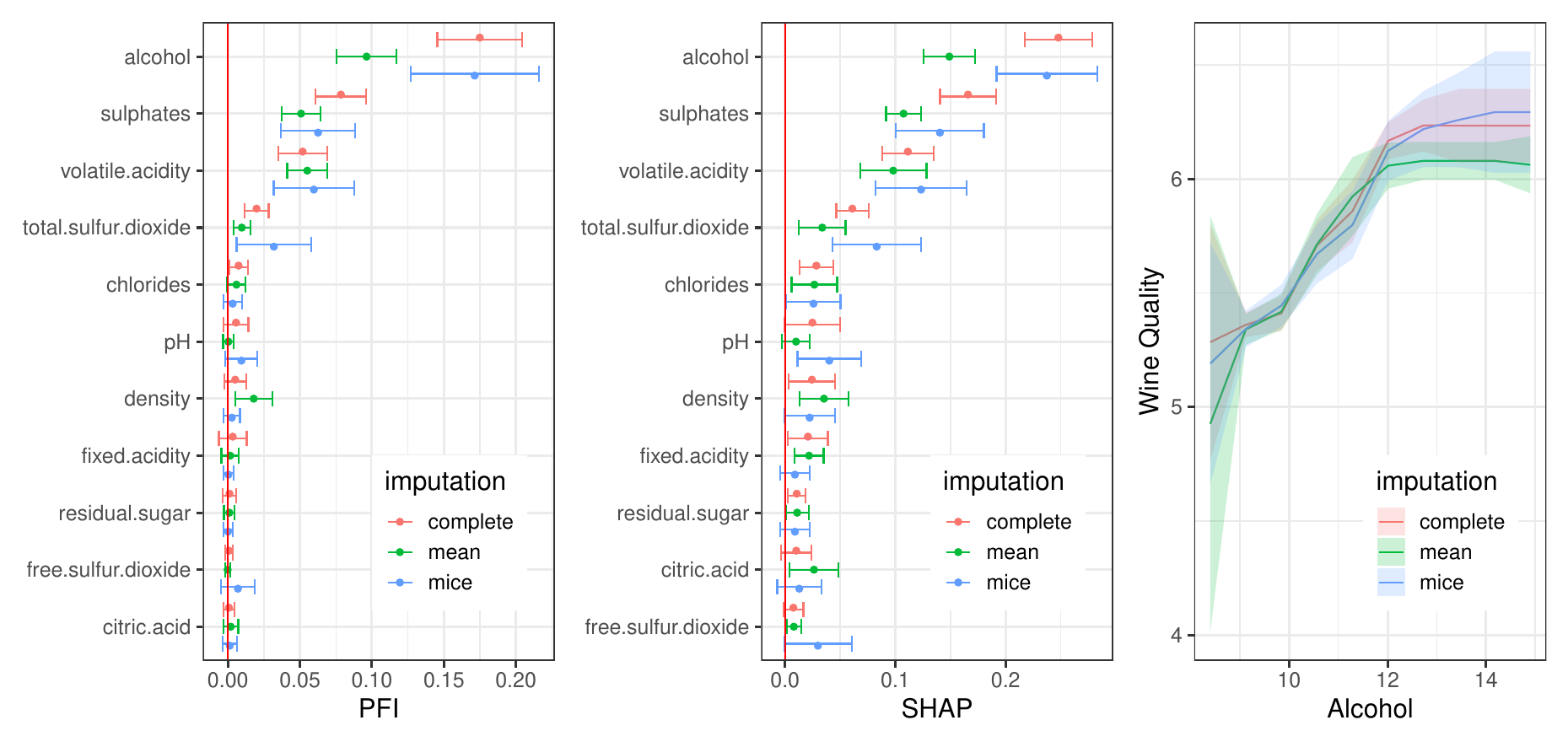}
    \caption{PFI (left), SHAP (middle) and PD plot (right) for XGBoost fitted on the wine dataset with adjusted bootstrapping with 40 \% simulated missing values according to MCAR. The model is fitted on the complete dataset (red) and on the imputed dataset by mean (green) and MICE (red) imputation.}
    \label{fig:real data}
\end{figure*}

Figure~\ref{fig:real data} shows the overall importance of the features according to PFI and SHAP and the effect of the most important feature \texttt{alcohol} in a PD plot.

According to PFI and SHAP, \texttt{alcohol}, \texttt{sulphates}, \texttt{volatile.acidity} and \texttt{total.sulfur.dioxide} are the most important features. While the order of importance is generally similar with both imputation methods, the importance of \texttt{alcohol} is drastically underestimated when mean imputation is used. Similar results, but less extreme, can be observed for the \texttt{sulphates} feature. Confidence intervals are generally wider with multiple imputation and narrower with mean imputation. 

For the PD plot, we notice that, on average, wine with higher \texttt{alcohol} content is of better quality, except for \texttt{alcohol} values above 12\%, where there is no further quality improvement. The PD with multiple imputation (MICE) is generally close to the PD on the full dataset. The PD with mean imputation is similar for \texttt{alcohol} values between 9\% and 12\% but shows a larger deviation for very low or high \texttt{alcohol} content. For most values, the confidence bands with MICE imputations are wider than those on the complete dataset, while those of the mean imputation are narrower. 

Results for all three IML methods show that mean imputation leads to underestimation of the effect of important features. This can be explained by the fact that more extreme missing feature values are replaced by the mean of the feature, leading to an underestimation of the feature's effect. For example, this can be seen in Figure~\ref{fig:real data}, where the PD is underestimated with mean imputation for large values of \texttt{alcohol}. The narrow confidence intervals with mean imputation show that it not only ignores imputation uncertainty but also reduces the width, compared to the complete dataset. That is because mean imputation drastically underestimates the variance. Multiple imputation, on the other hand, considers the imputation uncertainty in the variance estimation. This is reflected by the wider confidence intervals in Figure~\ref{fig:real data}. 

\section{Discussion} \label{sec: Discussion}
In summary, in our experiments, MICE PMM provides the most accurate interpretation for linear DGPs and MICE RF for non-linear DGPs.
However, we find that MICE often has a higher coverage than the model with complete data, which can be explained by the fact that MICE overestimates the variance. Therefore, when combined with bootstrap, which underestimates the variance, these two effects cancel out and often lead to the best coverage in our experiments. However, in practice, the two effects might not cancel out and could, in turn, lead to over- or underestimation of variance. 
%
%
%

For the average CI width, we find that it generally grows with the missingness rate. This is to be expected because the uncertainty grows with the missingness proportion. MICE methods have the largest CI width because they consider additional imputation uncertainty. Mean imputation, on the other hand, even leads to a decrease in variance because missing values are replaced by the mean of a variable. This explains the extremely poor coverage of mean imputation in these cases.
%

The positive bias for the imputation methods, except for MissForest, indicates that the importance is underestimated, while for MissForest it is overestimated. The underestimation can be explained by missing information and is what we expect for an analysis with missing data. The overestimation of MissForest might be explained by MissForest overfitting to the important features, since the target $y$ is included in the imputation. This shows that, while including the target in the imputation is important for valid inference with multiple imputation~\cite{white2011multiple}, it might lead to problems for MissForest.

\section{Conclusion}
We investigated how different imputation methods and the application of single or multiple imputation affect coverage and confidence intervals of IML methods. In simulation studies, we found that the application of multiple imputation provides good coverage, whereas MICE PMM performs better for linear DGPs and MICE RF better for non-linear DGPs. In the real data example, we demonstrated how much the importance can change for PFI, PD and SHAP, and how considering imputation uncertainty can improve the interpretation. Therefore, we recommend taking the imputation uncertainty into account using multiple imputation methods when interpreting ML models trained on data with missing values.


\section*{Acknowledgments}
PG and MNW were supported by the German Research Foundation (DFG), Grant Number 437611051. PG was supported by a PhD grant
of the Minds, Media, Machines Integrated Graduate School Bremen.

\bibliographystyle{unsrt}
\bibliography{bib_iml_unc.bib}


\clearpage
\appendix
\label{ap}

\begin{figure}[!htbp]
    \centering
    \includegraphics[width=0.92\linewidth]{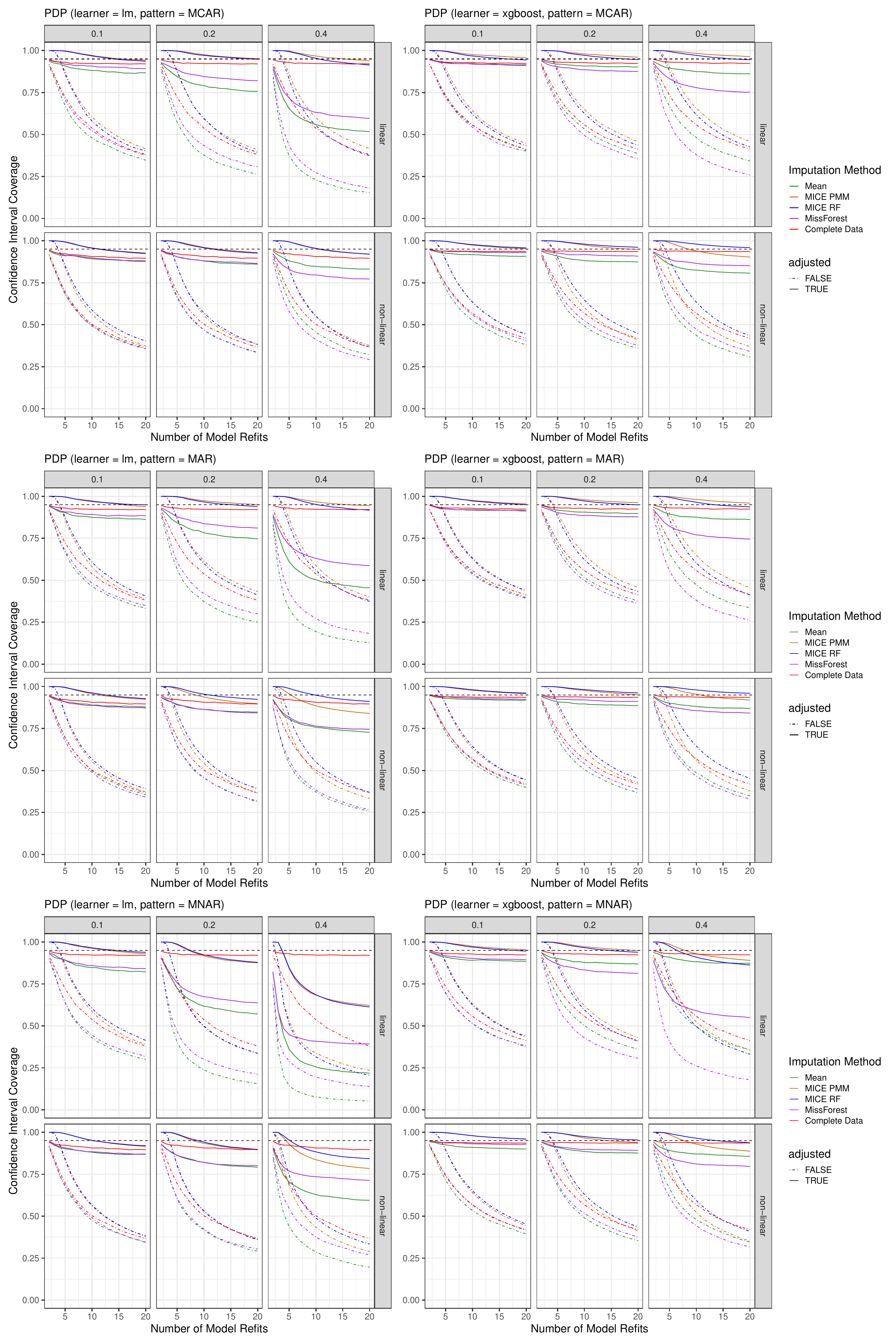}
    \caption{Coverage for PDP with bootstrap approach over model refits.}
    \label{fig: cov pdp}
\end{figure}

\begin{figure}[!h]
    \centering
    \includegraphics[width=0.92\textwidth]{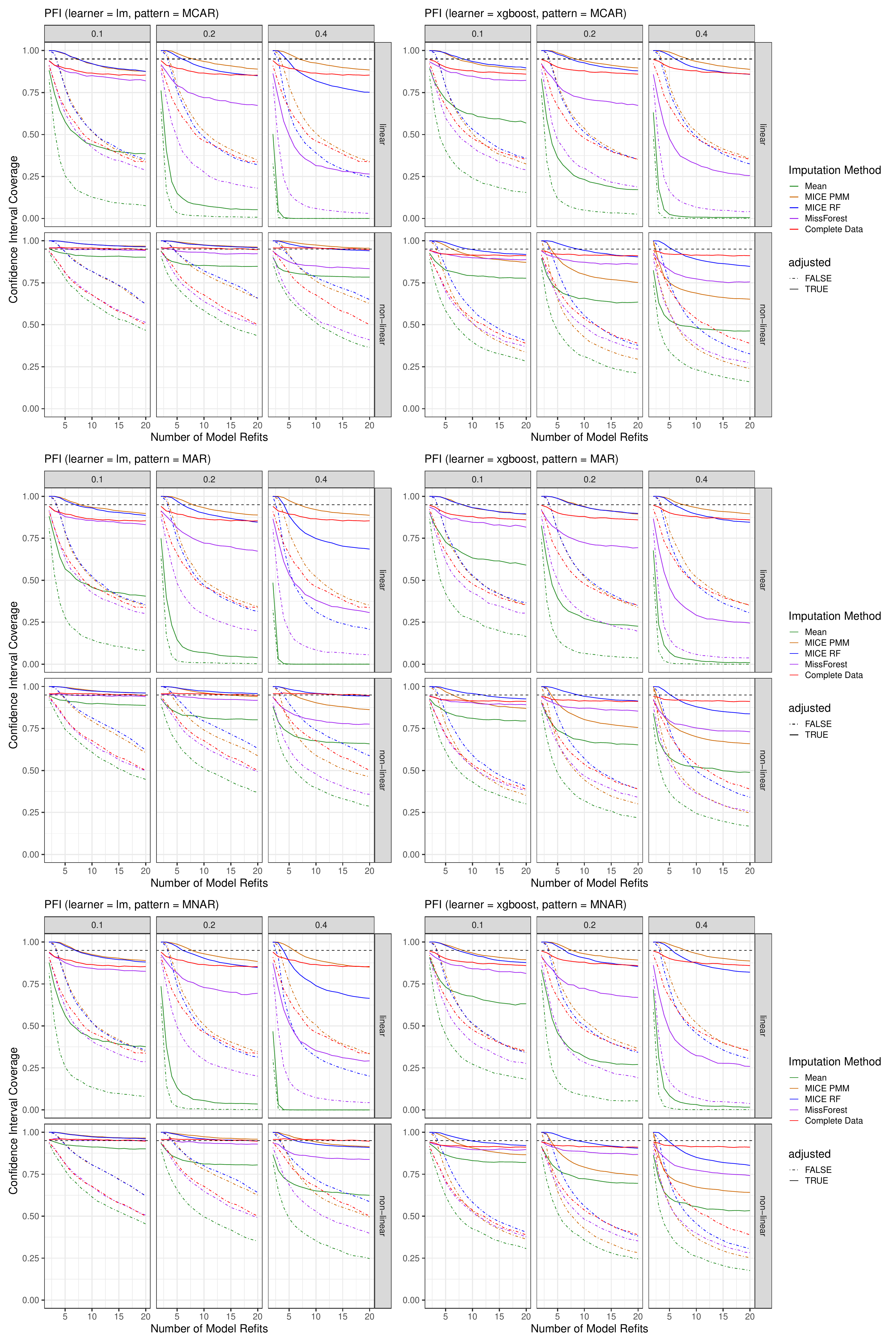}
    \caption{Coverage for PFI with bootstrap approach over model refits.}
    \label{fig: cov pfi}
\end{figure}

\begin{figure}[!h]
    \centering
    \includegraphics[width=0.92\textwidth]{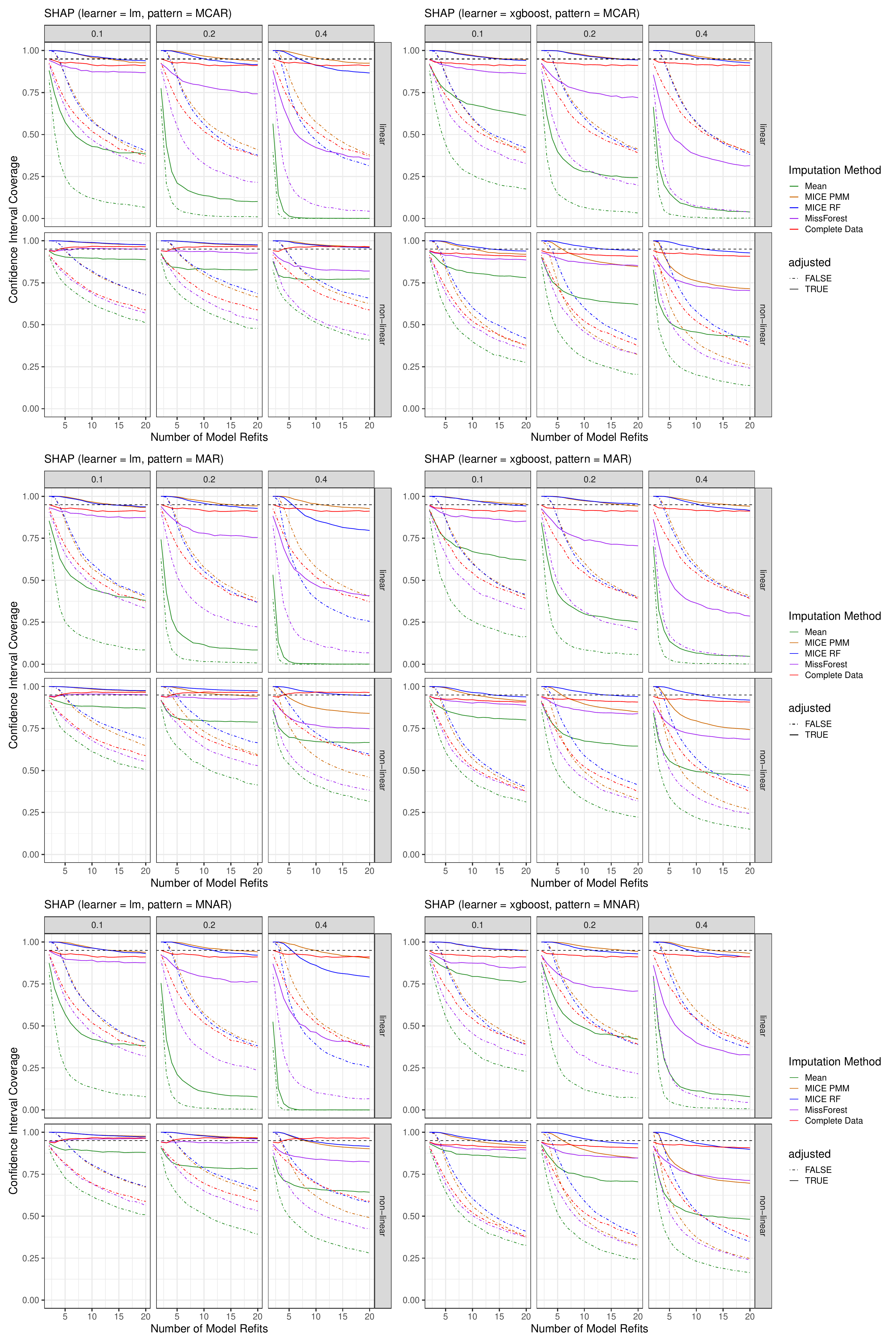}
    \caption{Coverage for SHAP with bootstrap approach over model refits.}
    \label{fig: cov shap}
\end{figure}

\begin{figure}[!h]
    \centering
    \includegraphics[width=0.92\textwidth]{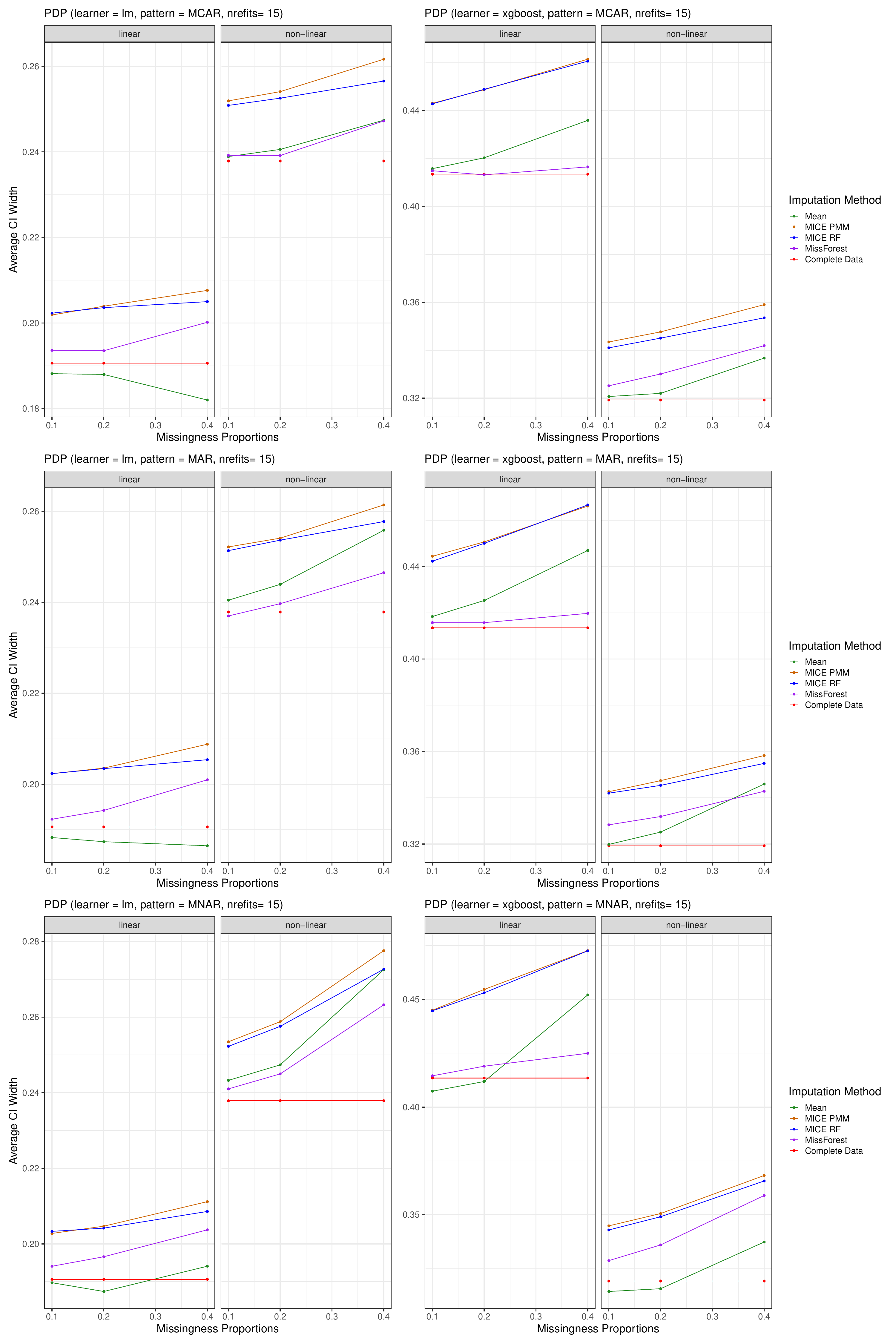}
    \caption{Average CI width for PDP with bootstrap approach across missingness rates for 15 model refits.}
    \label{fig: aw pdp}
\end{figure}

\begin{figure}[!h]
    \centering
    \includegraphics[width=0.92\textwidth]{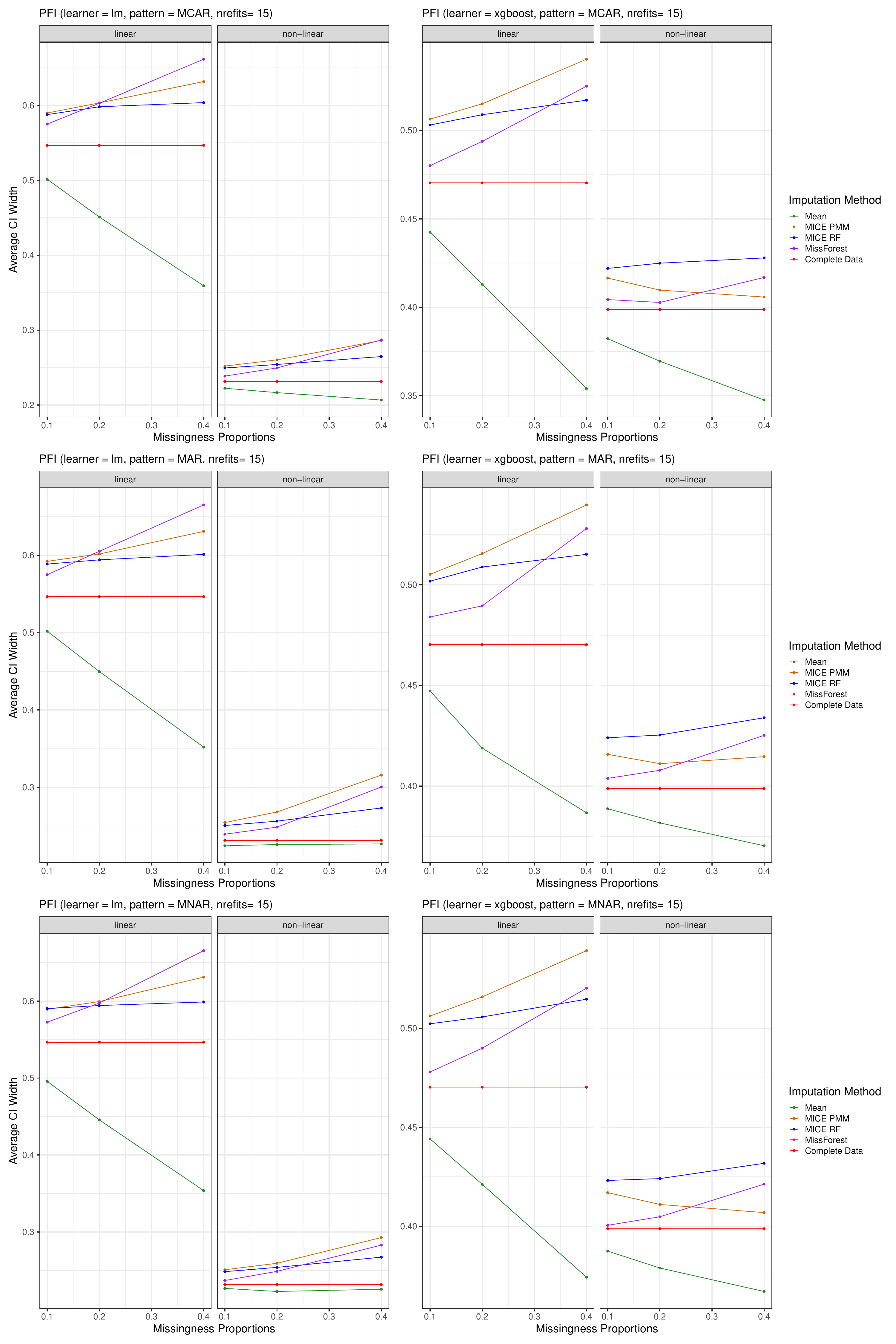}
    \caption{Average CI width for PFI with bootstrap approach across missingness rates for 15 model refits.}
    \label{fig: aw pfi}
\end{figure}

\begin{figure}[!h]
    \centering
    \includegraphics[width=0.92\textwidth]{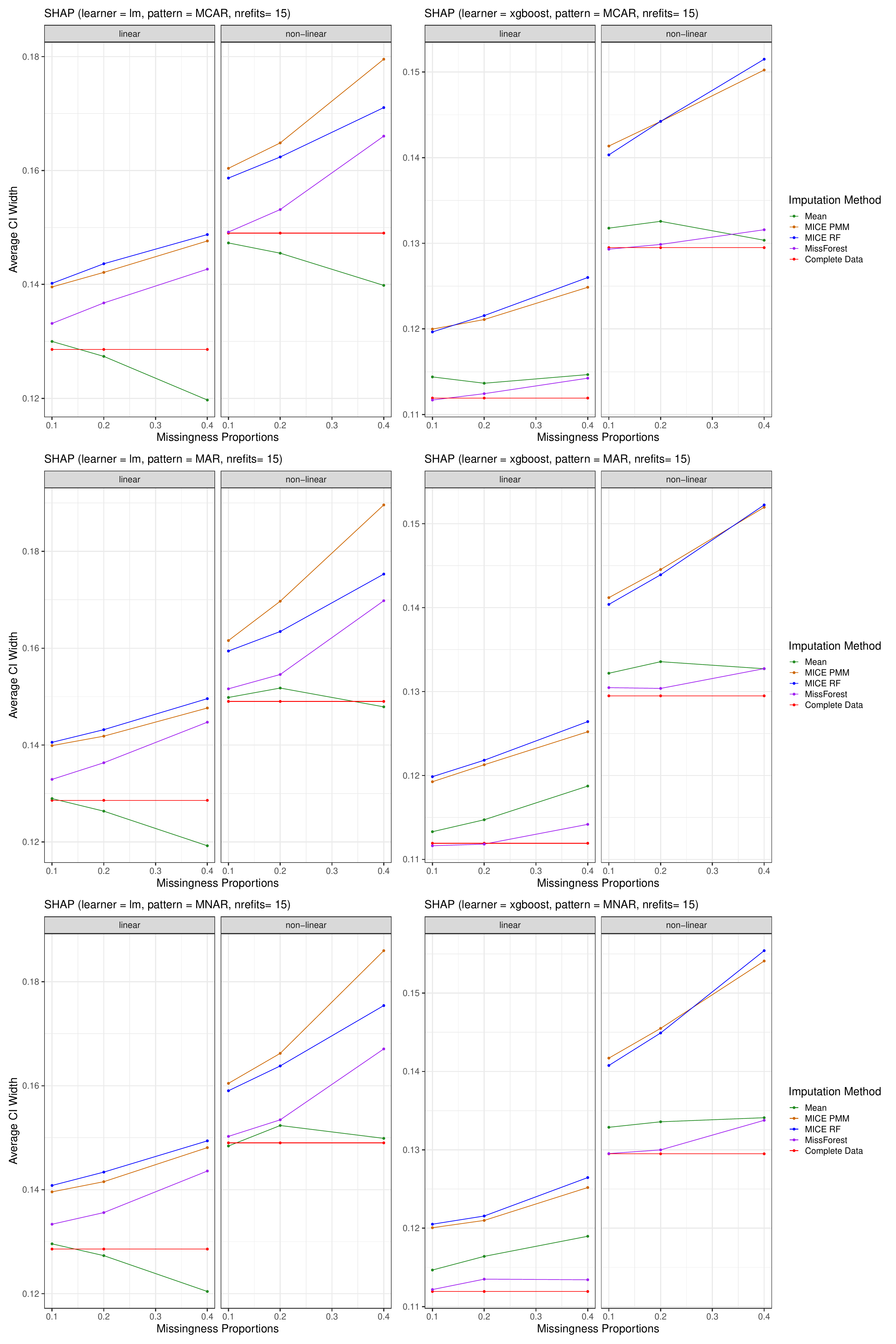}
    \caption{Average CI width for SHAP with bootstrap approach across missingness rates for 15 model refits.}
    \label{fig: aw shap}
\end{figure}


\begin{figure}[!h]
    \centering
    \includegraphics[width=0.92\textwidth]{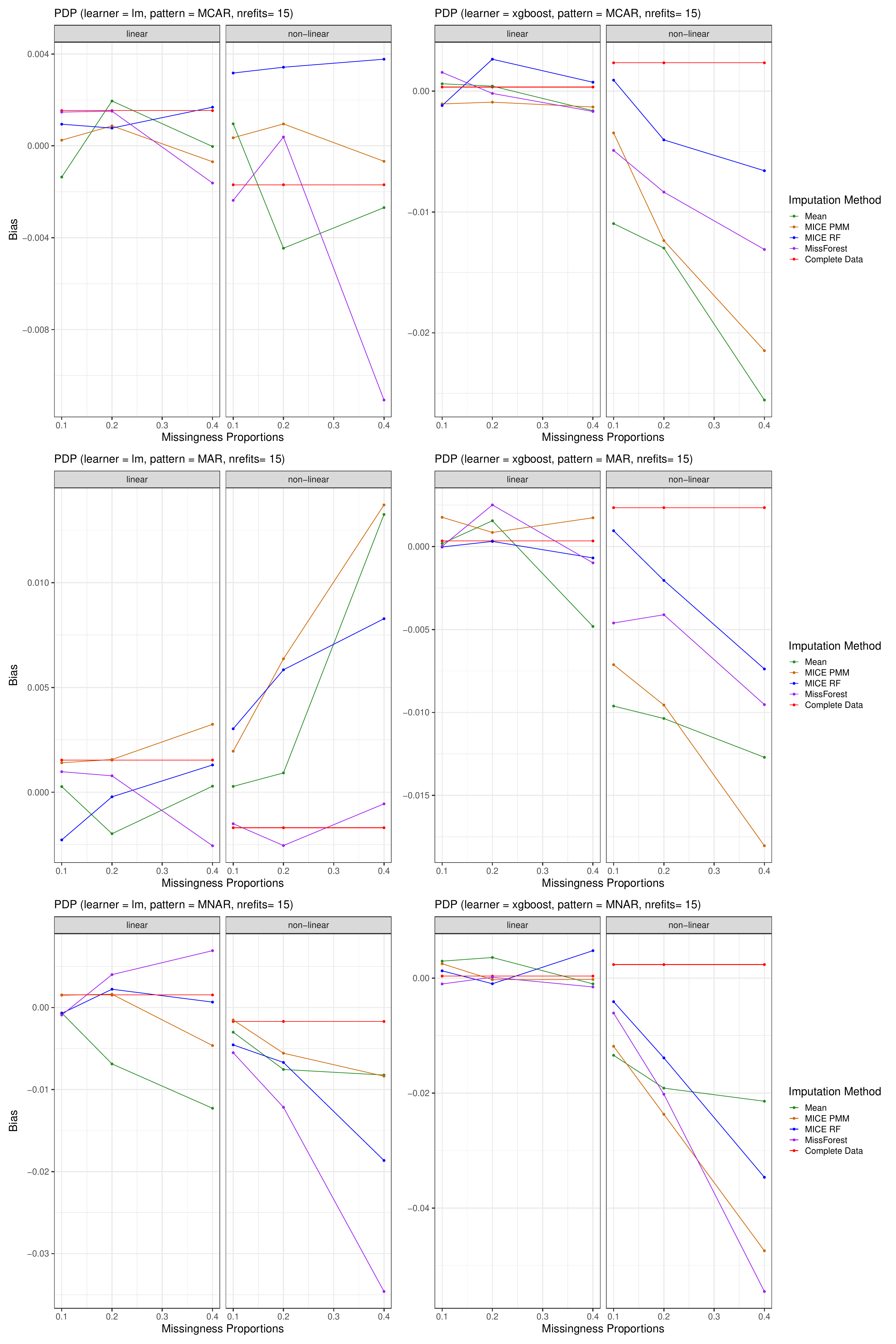}
    \caption{Bias for PDP with bootstrap approach across missingness rates for 15 model refits.}
    \label{fig: bias pdp}
\end{figure}

\begin{figure}[!h]
    \centering
    \includegraphics[width=0.92\textwidth]{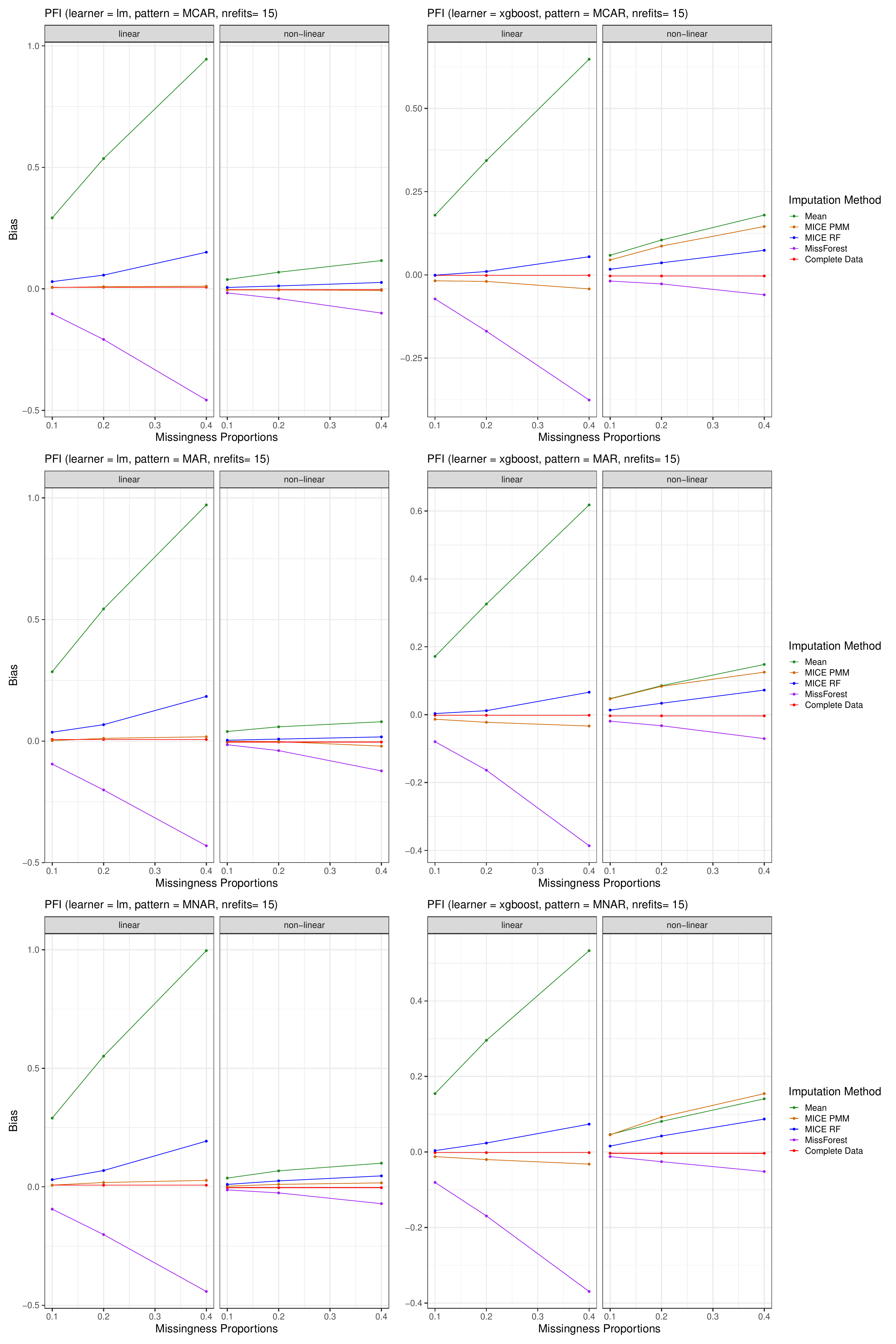}
    \caption{Bias for PFI with bootstrap approach across missingness rates for 15 model refits.}
    \label{fig: bias pfi}
\end{figure}

\begin{figure}[!h]
    \centering
    \includegraphics[width=0.92\textwidth]{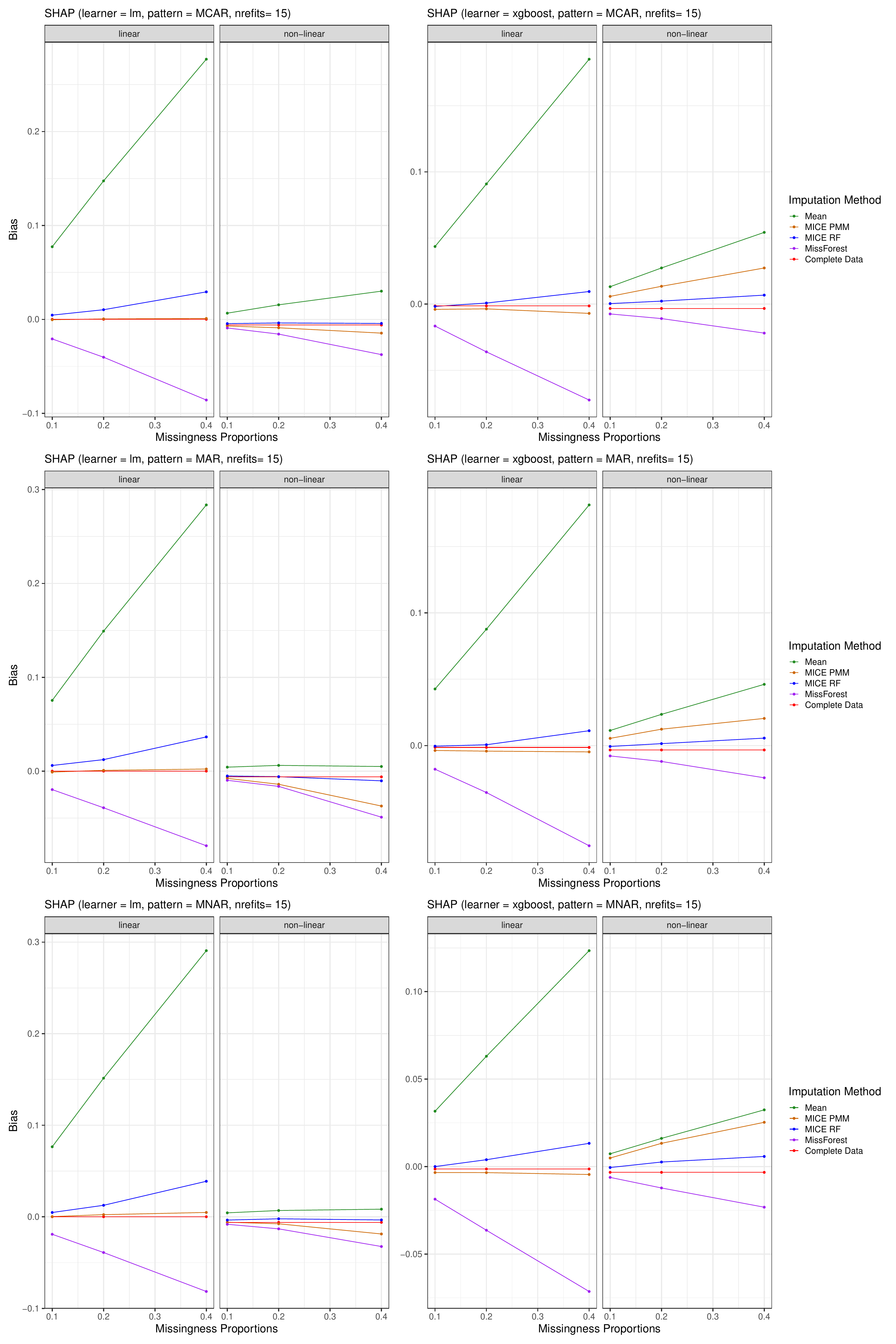}
    \caption{Bias for SHAP with bootstrap approach across missingness rates for 15 model refits.}
    \label{fig: bias shap}
\end{figure}

\end{document}